\documentclass{article}
\usepackage{hyperref}
\hypersetup{hidelinks}
\usepackage{ijcai25}
\usepackage{times}
\usepackage{soul}
\usepackage{url}
\usepackage[utf8]{inputenc}
\usepackage[small]{caption}
\usepackage{graphicx}
\usepackage{amsmath}
\usepackage{amsthm}
\usepackage{booktabs}
\usepackage{algorithm}
\usepackage{algorithmic}
\usepackage[switch]{lineno}
\usepackage{enumitem}
\usepackage{comment}
\usepackage{subcaption}
\usepackage{multirow}

\usepackage[table,xcdraw,dvipsnames]{xcolor}

\title{Explainable Reinforcement Learning Agents Using World Models}

\author{
Madhuri Singh
\and
Amal Alabdulkarim\and
Gennie Mansi\And
Mark O. Riedl\\
\affiliations
School of Interactive Computing, Georgia Institute of Technology.\\
\emails
\{msingh365, amal, gennie.mansi, riedl\}@gatech.edu,
}
\begin{document}
\maketitle

\begin{abstract}

Explainable AI (XAI) systems have been proposed to help people understand how AI systems produce outputs and behaviors. Explainable Reinforcement Learning (XRL) has an added complexity due to the temporal nature of sequential decision-making. Further, non-AI experts do not necessarily have the ability to alter an agent or its policy. We introduce a technique for using World Models to generate explanations for Model-Based Deep RL agents. World Models predict how the world will change when actions are performed, allowing for the generation of counterfactual trajectories. However, identifying what a user wanted the agent to do is not enough to understand why the agent did something else. We augment Model-Based RL agents with a Reverse World Model, which predicts what the state of the world should have been for the agent to prefer a given counterfactual action. We show that explanations that show users what the world should have been like significantly increase their understanding of the agent's policy. We hypothesize that our explanations can help users learn how to control the agent's execution through manipulating the environment.

\end{abstract}

\begin{figure}[t]
    \centering
    \includegraphics[width=0.86\columnwidth]{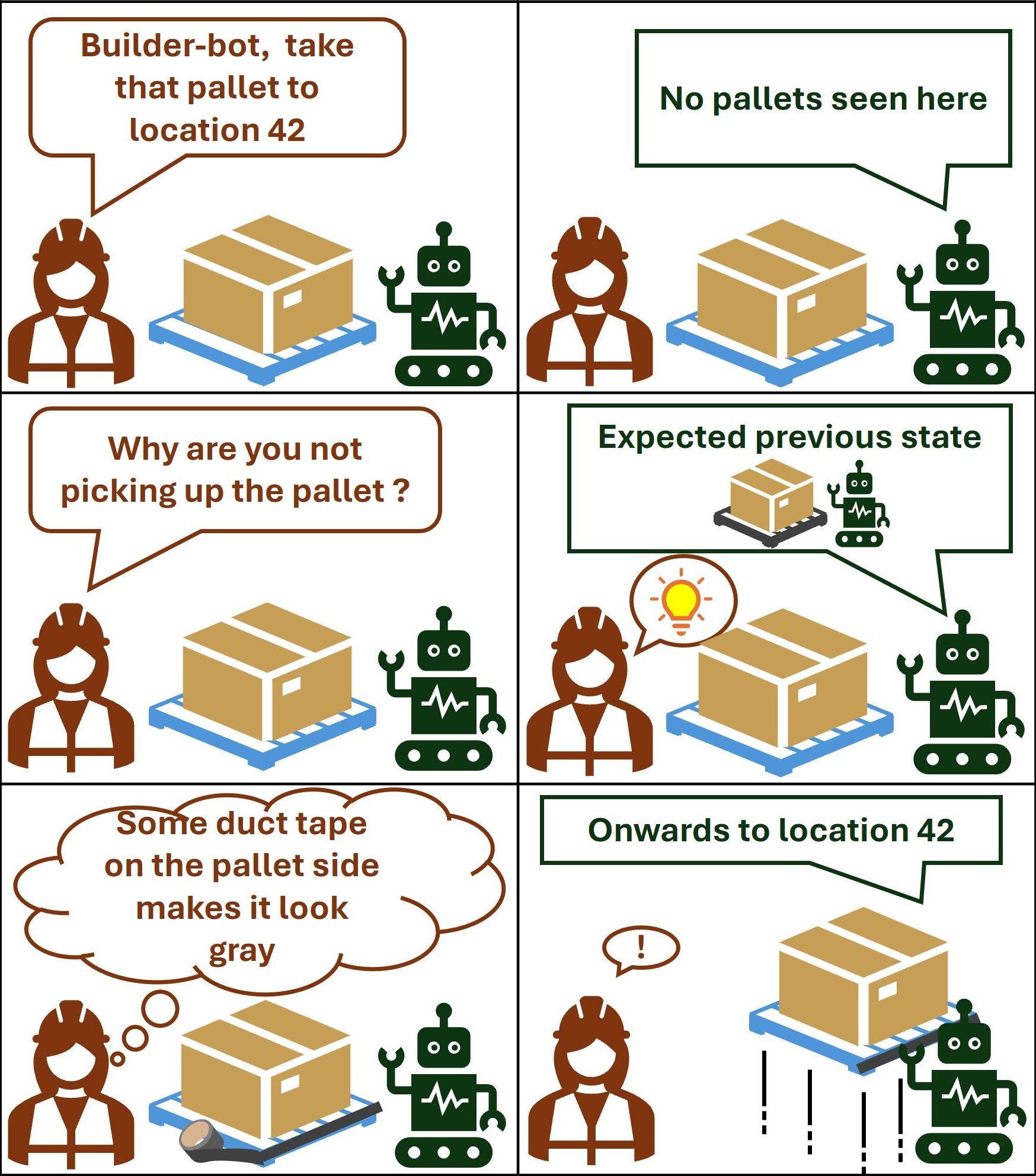}
    \caption{Image depicting how an agent can help a user understand why it is not performing actions to expectations, and how the user might change the world to induce the desired behavior from the agent.
    In this hypothetical scenario, the user may not have understood the significance of the color of the pallet to the agent. 
    The explanation demonstrates what the environment should have been like for the agent's policy to execute the desired action. 
    Armed with this new knowledge, the user is able to alter the environment to affect the desired behavior.}
    \label{fig:teaser}
\end{figure}

\section{Introduction}

Explainable AI (XAI) systems have been proposed to help people understand how AI systems produce outputs and behaviors.
Deep Reinforcement Learning (DRL) techniques learn a neural network policy model, which attempts to predict the action that is most likely to lead to future reward.
Because sequential, long-term behavior is encoded into the neural policy model, DRL agents are notoriously hard to debug and correct if their execution behavior diverges from user preferences, desires, or expectations.

If an XAI system can help users understand how the agent is responding to the environment, how the policy is making predictions, or how the policy was learned, they may be able to adjust to the agent or the environment, so that the agent's behavior aligns with our preferences for the policy.
Because DRL is applicable to agents and robot planning, many real-world applications, from self-driving cars to personal robotics at home, will have systems driven by DRL interacting with {\em users without technical background}. 
Non-AI experts will seek explanations for actions that deviate from their understanding of what the agent should be doing or how it should be doing it \cite{miller2019explanation}. 
Without an explanation, users can find it difficult to trust the agent’s ability to act safely and reasonably \cite{zelvelder2021assessing}, especially if they have limited artificial intelligence expertise. Even in the case where the agent is operating optimally and without failure, the agent’s optimal behavior may not match the user’s expectations, resulting in confusion and lack of trust. 
Non-AI experts may also wonder what they can do to influence or change the behavior of the agent if they cannot alter the internal workings of the agent or retrain the policy.

Explanation of reinforcement learning systems is especially challenging due to the temporal nature of decision-making---any given decision to perform an action may depend on future expectations as much as, if not more than, current observations.
This becomes more complicated when one accounts for users who are not AI experts and, furthermore, do not have the ability to change the algorithm or retrain the agent.
What would an {\em actionable} explanation look like for non-AI experts?

In this paper, we present a technique for using a World Model (WM) to provide actionable information to non-AI experts of DRL systems.
A WM describes the agent's understanding of the state-transition dynamics of the environment. 
In Model-Based DRL\cite{kaelbling1996reinforcement}, the agent learns a World Model through trial-and-error interactions with the environment during training.
The World Model is used to predict the effects of the agent's actions, often speeding up policy learning when interactions with the environment are slow or resource intensive.
In the scenario in which the agent fails a task or performs an unexpected action,
we show that the World Model can be used to generate a counterfactual explanation that shows the user what the agent would have expected the world state to look like in order to have chosen an action preferred by the user.
\textit{The user can then act upon this information to influence the agent's actions in the future by affecting change on the environment.}
This is because the agent's policy models the relationship between it and the environment with respect to action, and the WM makes that relationship explicit.
While the user may not be able to change or re-train the policy, the user can alter the agent's behavior by directly manipulating the environment if the agent has a robust policy.

\begin{figure}
    \centering
    \includegraphics[width=1\linewidth]{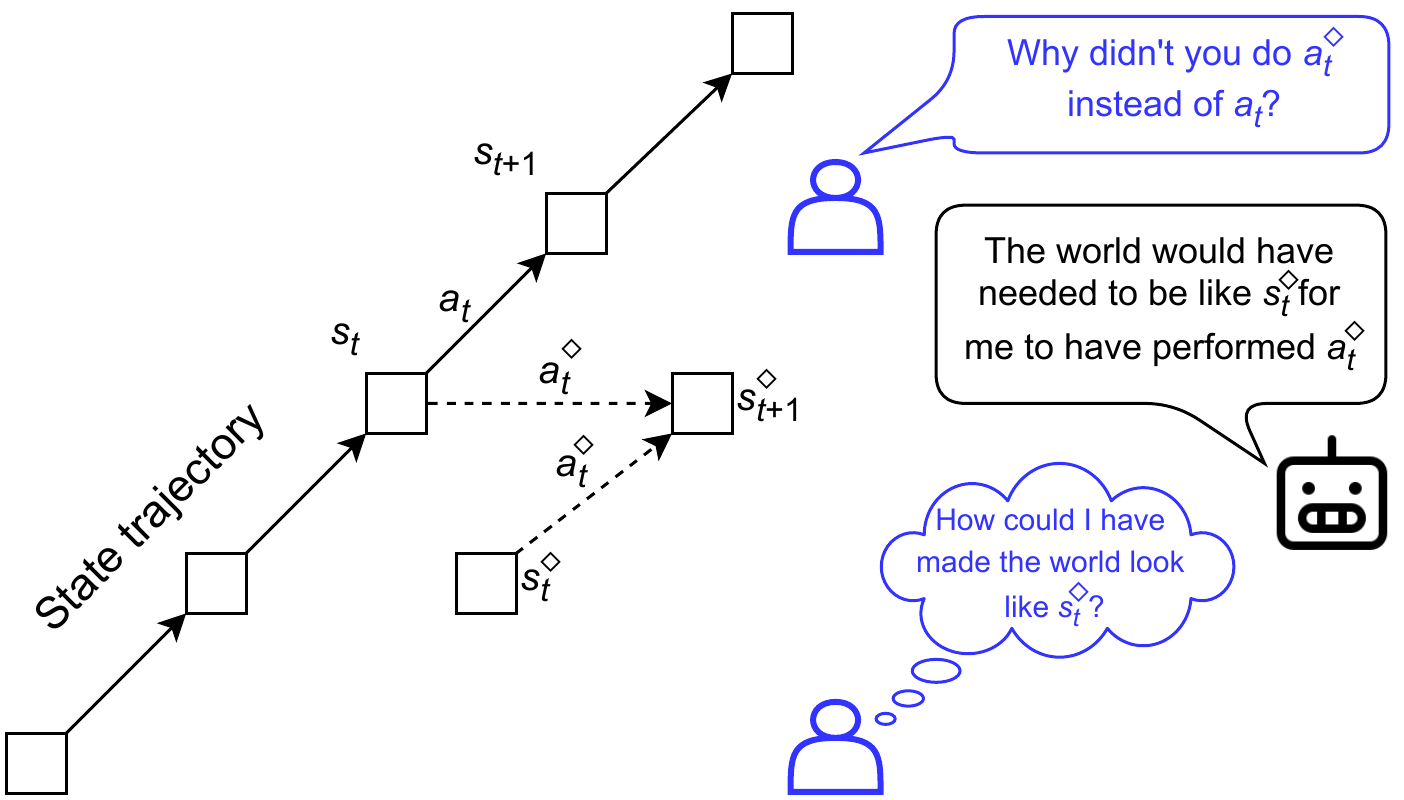}
    \caption{Diagram depicting an agent's trajectory through state-action space. In state $s_t$, the agent's policy preferred action $a_t$, causing a transition to $s_{t+1}$. The user wanted the agent to pick action $a^\diamond_t$. The forward world model can tell us that the effect of $a^\diamond_t$ is state $s^\diamond_{t+1}$. A reverse world model can generate $s^\diamond_t$, the state that the agent should have been at time $t$ for its policy to pick the desired  $a^\diamond_t$. } 
    \label{fig:state-diagram}
\end{figure}

World Models used for a reinforcement learning agent to learn its policy only need to predict a probability distribution over states that can follow a given state and a given action, $Pr(s_{t+1}|s_t, a_t)$.
We refer to this as the {\em forward world model}. 
The forward world model can generate a counterfactual trajectory corresponding to what the user might have indicated that they expected. 
That is, if we give it an alternative action $a^\diamond_t$, the WM will tell us what would have happened instead, $s^{\diamond} _{t+1}$.
What would the environment need to have been like for the agent to pick $a^\diamond_{t}$?
Unfortunately, it is not the state the agent was actually in, $s_t$.
We must identify some counterfactual state, $s^\diamond_t$, that the agent was {\em not} in at time $t$ that would have induced the agent to prefer $a^\diamond_{t}$ over $a_t$.
To generate $s^\diamond_{t+1}$, we require a {\em reverse world model} that generates $Pr(s_t | s_{t+1}, a_t)$, the distribution over states that we should have been in at time $t$ to pick the action $a^\diamond_t$ that would have delivered us to desired state $s^\diamond_{t+1}$.
See Figure~\ref{fig:state-diagram}.

The predicted counterfactual state $s^\diamond_t$ can be presented to the user as an explanation of what state the world and agent needed to be in for the agent's policy to have performed the expected/desired action.
Using a virtual agent performing household tasks,
We show that presentation of these states to non-AI experts improves user  understanding of why an agent fails to perform as expected and what would need to change in the environment for the agent to succeed.
We further show that our explanations increase user satisfaction, increase user trust, and decrease user cognitive load when trying to identify why the agent failed.

\section{Related Work}

Explainable Reinforcement Learning (XRL) methods can be classified based on which aspect of the RL agent they explain, which can be at the Policy Level (PL), Feature importance Level (FI), or the learning and Markov Decision Process (LPM) \cite{Milani_2024}. FI methods focus on providing the immediate context for single actions. Directly generated natural language explanation methods, such as rationale-generation \cite{ehsan2019automated}, are an example of FI explanations. LPM methods provide additional information on the effects of the MDP or training process, such as influential training experiences or how the agent acts regarding its rewards or objectives. Often, LPM methods require additional information, such as concepts in State2Explanation \cite{das2023state2explanation}, graphical causal models \cite{madumal2020explainable,peng2022inherently}, a learned transition model \cite{van2018contrastive}, and partial symbolic model approximations that capture actions and their preconditions \cite{sreedharan2022bridging}, as well as reward decompositions \cite{alabdulkarim2025experiential,das2020leveraging,septon2023integrating}. Lastly, PL explanations summarize long-term behaviors using abstractions or representative examples. Agent Strategy Summarization \cite{amir2018agent} is a representative example of a PL method. Our method is an LPM method that models domain information by learning with the agent’s world model.

Counterfactual explanations, in particular, 
provide high-level, actionable information which is suitable for non-AI expert users \cite{10.1145/3648472}. 
People naturally explain in ``the form `\textbf{Why P rather than Q?}', in which P is the target event and Q is a counterfactual contrast case that did not occur" ~\cite{miller2019explanation}. 
Counterfactual (CF) explanations answer the question, 
what needs to be changed in the input of a computer system such that its output changes to a different one. 
In sequential decision-making paradigms the output is the choice of a different action than the one the model has taken or wants to take next.
In sequential decision-making, counterfactuals can also take the form of hypothetical future action trajectories to show a user what future expected states could be achieved if an alternative action is taken.

Most closely related to our work are CF explanation systems that generate counterfactual states for which the agent would have chosen a different, desired action if those states were encountered~\cite{OLSON2021103455,10.5555/3545946.3598751,10.1145/3709146}.
These works employ a second round of training to create a separate, explanation generation model, using action trace data produced by a pre-trained RL agent.
This approach has been noted to potentially create counterfactual states that contain the requisite surface features necessary to push the agent to select a new action, but does not guarantee that those states are reachable in the state space~\cite{10.5555/3635637.3662915}.
Our technique instead generates explanations from the world model, which is trained as part of the RL agent's policy training loop.
Our technique does not require a separate post-hoc training phase, and the world model---a predictor of successor and predecessor states---is more likely to produce traversable states. 

The RACCER system~\cite{10.5555/3635637.3662915} generates counterfactual states through a search of reachable states for those with the desired properties.
It requires access to the execution environment---a simulation, or real-world---to conduct the search.
Our technique does not require access to the execution environment, as the world model is a learned surrogate for the environment's transition dynamics.

COViz~\cite{amitai2024cfactionoutcomes} takes a different approach to counterfactuals and generates states that are predicted to be visited if an alternative action were to be chosen over the actually chosen action.
This is potentially useful information to users in evaluating whether an alternative is better than an actual action choice.
This approach is complementary to our work, which helps users determine how to induce the agent to take a different action without explicit override.

\section{Generating Actionable Explanations}

A critical question in XAI is what an {\em actionable} explanation would be for reinforcement learning.
Actionability refers to how information in explanations helps users take actions in response to an underlying AI system\cite{mansi2023dontitoutliningconnections}. For an AI system developer, actionable explanations about a RL system often center actions that involve changing the policy, such as correcting the algorithm or altering how the AI system is trained. Non-AI expert end-users cannot change the policy, but explanations for RL systems can still be actionable.  

Researchers~\cite{alabdulkarim2025experiential,chakraborti2021emerging,das2023state2explanation}
have proposed that explainable RL should update the user's understanding of how the agent responds to the environment. 
By changing users' mental model of the agent, explanations can also help users understand how they can change {\em their own behavior}. For example,  policies are learned responses to the local environmental state respective to task reward. Consequently, explanations can be actionable by helping users understand how they can change the environment, so the RL system responds as they wish.  

In order to enable this kind of actionability, it is critical for users to have a mental model of how changes in the environment influence the agent's behavior. Our method provides this information by showing users what the environment should have looked like for the agent to take an alternate path. This can allow users to physically alter the environment in order to gain control over an agent, helping users respond to an agent that is not executing as expected or desired.

We hypothesize that non-AI expert end-users should, with our explanations, be able to identify the root cause of what features of local environmental observations resulted in an AI agent performing an unexpected or undesirable action. We test this hypothesis in Section~\ref{sec:study}. 
In constructing this understanding, we further speculate that users that receive our explanations can more easily reason about what they could do to adjust the agent's environment to induce the desired behavior in the future.

\section{Preliminaries: Model-Based RL}

Before we present our world modeling technique, we review reinforcement learning and the specific version of model-based RL that we build upon.

Reinforcement learning solves sequential decision-making problems. 
Specifically, it generates a policy, $\pi(s)\rightarrow a$, that maps a state observation $s$ to an action $a$ such that executing the action optimizes future expected reward if the policy is followed henceforth.
Deep reinforcement learning approximates the policy with a neural network.
The policy model is charged with interpreting the state observation with respect to future expected reward.
The policy thus captures the dynamics of the agent with respect to the environment and task.

Model-based reinforcement learning is an RL approach that involves learning an environment model and then using it to train the policy \cite{kaelbling1996reinforcement}. 
The World Model \cite{Ha2018-lp}, also referred to as the transition dynamics model, approximates the distribution over next state observations given a state and an action: $Pr(s_{t+1} | s, a)$.
While a world model is not strictly required for reinforcement learning, in many cases, it reduces the number of interactions an agent must conduct in the actual environment because the agent can simulate world state transitions with its world model. 

The DreamerV3~\cite{hafner2025mastering} framework is one of the more successful model-based DRL frameworks. 
It incrementally learns a world model through interactions with the environment and trains the policy against the world model.
DreamerV3 comprises of the following models:

\begin{equation}
\scriptsize
\label{eq:dv3}
\text{DreamerV3: } 
\begin{cases}
\text{Sequence model: } 
h_t = f_{\phi}(h_{t-1},z_{t-1},a_{t-1})\\
\text{Encoder: }
Pr_{\phi}(z_t | h_t, x_t)\\
\text{Dynamics Predictor: }
Pr_{\phi}(\hat{z_t}|h_t)\\
\text{Decoder: }
Pr_{\phi}(\hat{x_t}|h_t,z_t)\\
\text{Reward predictor: }
Pr_{\phi}(\hat{r_t}|h_t,z_t)\\
\text{Continue predictor: }
Pr_{\phi}(\hat{c_t}|h_t,z_t)
\end{cases}
\end{equation}
where $\phi$ describes the parameter vector for all distributions optimized, and
\begin{itemize}[noitemsep,topsep=0pt]
    \item $x_t$ is the current image observation.
    \item $h_t$ is the encoded history of the agent.
    \item $z_t$ is an encoding of the current image $x_t$ that incorporates the learned dynamics of the world.
    \item $s_t = (h_t,z_t)$ is the agent's compact model state.
\end{itemize}
Of particular relevance for this work is 
the dynamics predictor model, $p_{\phi}(\hat{z_t}|h_t)$, which predicts an image encoding $\hat{z_t}$ given the encoded history of the agent, $h_t$. 
For efficiency the DreamerV3 world model operates in the latent embedded state space.
Predicted state observations $\hat{x_t}$ can be generated via the image prediction model $Pr_{\phi}(\hat{x_t}|h_t,z_t)$.

\section{Reverse World Models}

Model-based RL, and DreamerV3 in particular, learns to predict the (embedded) next state because it is seeking to understand how actions result in {\em future} expected reward.
While the world model is learned primarily to optimize training efficiency with respect to policy performance, we observe that the world model itself can be used to generate counterfactuals---how would the world change if a different action were to be taken. 
We operate in the setting where the user gives us the counterfactual action, $a^\diamond_t$, and, by implication, the counterfactual next state $s^\diamond_{t+1}$.
However, the explanation is not what the user already knows or can infer.
Our explanation generation strategy hinges on providing the user with an understanding of what the world {\em should have looked like} prior to $s_t$ for the agent to want to take action $a^\diamond_t$ instead of its actual chosen action $a_t$.
That is, the explanation is constructed around the presentation of $s^\diamond_t$, the world the agent should have been in to ``do the right thing''.
A forward world model cannot do this.  To generate $s^\diamond_t$ we need a {\em Reverse World Model}.

\begin{figure*}[t]
    \centering
    \includegraphics[width=0.8\linewidth]{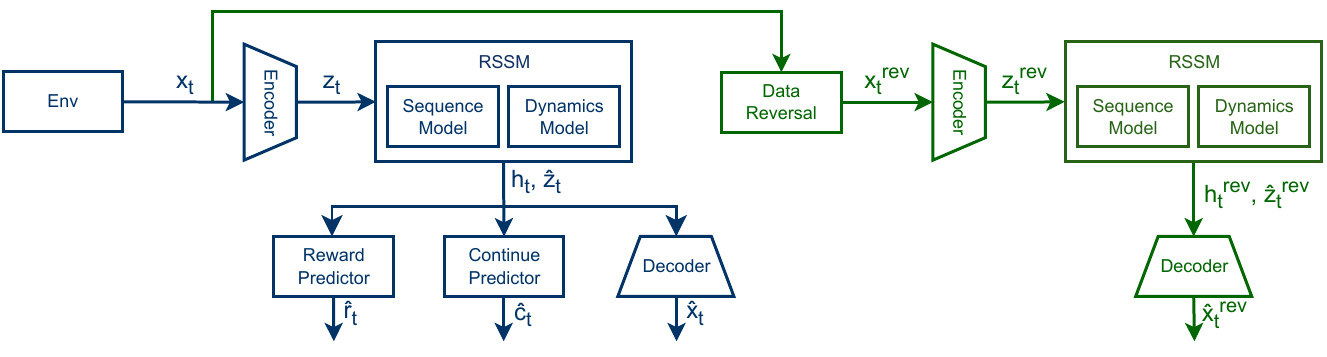}
    \caption{Our modified world model. The left-hand side is the forward world model, as in DreamerV3{\protect\cite{hafner2025mastering}}. The right-hand side shows the reverse world model.}
    \label{fig:simple_fwm_rwm_comparison}
\end{figure*}

The Reverse World Model (RWM) predicts the embedded state $p_\phi(\hat{z}_{t-1}|h_{t-1}) \text{, where } h_{t-1}\text{ is a function of }h_t, z_t\text{ and }a_{t-1}\text{,}$
$\text{ and }z_t \text{ is obtained from }x_t$
. The RWM trains alongside the WM, on a modified copy of the same training data, as follows.
First, reward and continuation
 data is removed.
Second, the temporal order of the state-action transition data generated during trials is reversed.
DreamerV3 trains its models using 
an experience replay buffer.
DreamerV3 samples chunks of transition data $(x_t, a_t)|_{t=k...k+n}$ from the replay buffer.
When training the RWM only, we simply reverse the order of the data sampled from the replay buffer.
Finally, we shift the actions so that image $x_{t+1}$ and the {\em prior} action $a_t$ are paired so that when the RWM is trained it predicts the prior embedded state given the future state and prior action.

Once the data processing is done, the RWM proceeds with training based on the usual dynamic and representation losses but with prediction loss only considering the decoder's image predictions.
The DreamerV3 Forward World Model with our combined Reverse World Model is shown in Figure~\ref{fig:simple_fwm_rwm_comparison}. 
The Reverse World Model produces $\hat{z}^\mathrm{rev}_t$, the latent encoding of the prior world state, and $\hat{x}^\mathrm{rev}_t$, the reconstructed image from the latent.
The RWM module is trained using reconstruction error between the environmental state observation $x_t^\mathrm{rev}$ and the image decoded from the latent, $\hat{x}_t^\mathrm{rev}$

\section{Human Participant Study} \label{sec:study}

Since explanations are meant to provide actionable information to non-AI experts, we conduct a human participant study.
We hypothesize that non-AI expert end-users should, with our explanations, be able to identify the root cause of what features of local environmental observations resulted in an AI agent performing an unexpected or undesirable action.
Specifically, we make the following hypotheses of those who receive explanations relative to those who do not receive an explanation:
\begin{enumerate}[label=H\arabic{enumi}.]
\item Participants that receive our explanations can more accurately recognize the causes of agent failures.
\item Participants that receive our explanations have greater satisfaction in agent responses.
\item Participants that receive our explanations have higher trust in the agent.
\item Participants that receive our explanations have lower cognitive load when identifying the causes of agent failure.
\end{enumerate}

We ran our human study as an online Qualtrics survey hosted on the Prolific platform. We recruited a total of 70 participants (Control Group strength = 33, Treatment Group strength = 37) in a randomized trial. 

The participants were in the age range of 19 to 73 (Mean: 41.3, Standard Deviation: 13.9), with 52.9\% people identifying as women and 47.1\% identifying as men. 
The study had a median run time of 21:03 minutes. We paid participants at the rate of \$12.00 per hour, with an added bonus of \$0.50 to be given for high-quality responses, which was given to all participants. 

\subsection{Study Design}

Participants are asked to imagine they are attempting to identify the root cause for why a robot (in a virtual game-like simulation) fails to correctly execute, or inefficiently executes, a task of making coffee.
Because we want to control for users' commonsense understanding, we present a fictional world in which coffee can be made with atypical ingredients like lava, or milk obtained straight from a cow. The recipes also change between the tasks, so the coffee may be made on a stove in one recipe and in a microwave in another.

\begin{figure}[t]
    \centering
    \includegraphics[width=0.6\columnwidth]{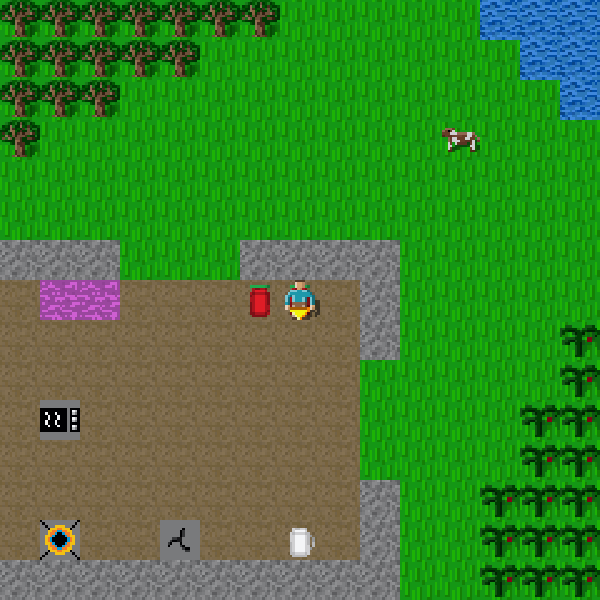}
    \caption{The modified Crafter environment. The agent must interact with various objects to gather ingredients and then interact with various devices to complete the recipe.}
    \label{fig:crafter}
\end{figure}

Participants watch a video of the agent performing its task in a virtual environment built using the Crafter \cite{hafner2022benchmarkingspectrumagentcapabilities} environment (See Figure~\ref{fig:crafter}). 
Crafter is an extensible 2D implementation of the MineCraft game.
We constructed a world consisting of a kitchen and fictional coffee ingredients that can be gathered from nearby environs.
The inventory mechanics are maintained, and we implement a new requirement that the agent complete a fictional coffee recipe.
Each scenario plays out in a $15\times 15$ grid and the agent can observe a $7\times 4$ window around its position and its inventory.

We present four scenarios, in each of which the agent fails to complete the fictional coffee recipe for a different reason (see description of scenarios below). 
For each scenario, the participant must identify what was tampered with in the environment that caused the agent to operate ineffectively.
In the Control condition, participants were presented with (a)~the video, and (b)~four snapshots of states during the agent's execution.
In the Treatment condition, participants were presented with the same information as above, but also
(c)~each snapshot was paired with a generated image showing what the agent was expecting in order to choose the action that would have resulted in correct execution. 

Hypothesis H1 holds if the explanation, in the form of the state expected prior to the counterfactual provides enough information for the participant to pick the reason why the agent chose the wrong action.
See Figure~\ref{fig:qualtrics-question} in the Appendix.

For the participants to indicate the cause of the failure, they select one of four possible objects from a pre-populated  list, and one of four verb-phrases from a pre-populated list. All objects and verb-phrases are probable---each object is present in at least one of the snapshots---but only one combination of object and verb-phrase is correct---there is a 1:16 chance of randomly selecting the correct combination.
See Figure~\ref{fig:qualtrics-answer} in the Appendix.

There was also a free-response text box where participants were asked to describe their reasoning for their choice.
This discourages participants from random selection but was not analyzed further.

\subsubsection{Scenarios}

The four scenarios involved completing the fictional coffee recipe, but under one of the following conditions:
\begin{enumerate}
\item A non-essential object is removed
\item An essential object is moved
\item An essential object is obstructed
\item An essential object is removed
\end{enumerate}
The order in which scenarios are presented to participants is randomized.

\subsubsection{Surveys}

After the four scenarios, participants completed three surveys: satisfaction, trust, and cognitive load.

The satisfaction survey is adapted from Hoffman et al.~\cite{Hoffman_2023}. 
It consists of seven 5-point Likert scale questions that ask about several key attributes of explanation satisfaction.
All questions are framed positively, and the answers are supposed to range from `Strongly Agree' to `Strongly Disagree'. We mapped these responses as `Strongly Agree'=5 down to `Strongly Disagree'=1 for simplifying later calculations. 
We edited the wording of the questions to fit our study while maintaining the same meaning. To have a fair baseline, both the groups were asked these questions about the Error Reports. The error reports presented to the different participant groups differed in that the treatment group received the extra RWM-generated snapshots. 

We used the User Trust Survey, also from Hoffman et al.~\cite{Hoffman_2023}. 
This survey consists of eight 5-point Likert scale questions. 
All questions are framed positively, except one (``I am wary of the tool'').

Our third survey is the NASA Task Load Index (TLX) Survey~\cite{hart1988development}, which assesses perceptions of cognitive load. It consists of six questions, each answered on a scale of 21 gradations.

\subsection{Agent Configuration}

The agent is our modified DreamerV3 model with a reverse world model. It is an Actor-Critic model trained alongside a 25 million parameter FWM and a 25 million parameter RWM. The actor and critic are both MLPs, with learning rates of 3e-5, a batch size of 8 and a batch length of 65. The observation input is an image of a 7x7 grid which depicts the agent's field of view at a given time step. The full training grid for each static environment is 15x15.

The agent never experiences items being moved, obstructed, or removed during training.
It is {\em intentionally} overfit for purposes of experimental control.
While overfit policies are generally unwanted, we required an unnaturally brittle agent that would struggle with the task to construct our experimental conditions; 
it makes the point that no agent in a complex real-world setting can be perfect.

The aim of this human study is to validate whether the RWM generated explanations are genuinely useful to people. To accomplish this, in summary, we trained four RL agents to perform a task in a static environment. We then introduced unexpected changes (e.g. items missing) to said environments and recorded the changed trajectories of the trained agents. We then showed this recording to the study participants along with the RWM-generated expectations of what the agents expected the environment to look like. We then asked these participants if they were able to correctly identify the change. 

The agents are the same between experimental and control conditions except for the presentation of the explanations generated by the reverse world model.
For each scenario, we selected the point at which the agent deviated from the optimal solution trajectory and three other random points and the snapshots are presented to participants.

\section{Results}

\subsection{Accuracy}

We looked at the percentage of participants in each group who got the answers right per scenario.
Table~\ref{tab:accuracy} shows the average correctness of participant selections. 
We used Fisher's Exact Test for categorical data to evaluate statistical significance. 
Participants in the explanation condition were significantly ($p<0.008$) more likely to identify the cause of the agent's problem and correctly assemble that reason using the  pre-populated  lists of objects and verb phrases.
{\em Hypothesis H1 is supported.}

\newcommand{\ra}[1]{\renewcommand{\arraystretch}{#1}}

\begin{table}[t]
\ra{1.3}
\resizebox{\columnwidth}{!}{
    \begin{tabular}{lrrrr}
        \toprule
        \multirow{2}{*}{\textbf{Scenario}} & \multicolumn{2}{c}{\textbf{\% Correct}} & \multirow{2}{*}{\textbf{Odds Ratio}} & \multirow{2}{*}{\textbf{p-value}}\\ 
        \cmidrule{2-3}
        & \textbf{Control} & \textbf{Treatment} & &\\
        \midrule
        Remove Non-Essential Object & 3.03 & 27.03 & 11.85 & 0.00571* \\ 
        Move Essential Object & 33.33 & 86.49 & 12.80 & 0.00001* \\ 
        Obstruct Essential Object & 27.27 & 62.16 & 4.38 & 0.00337* \\ 
        Remove Essential Object & 42.42 & 83.78 & 7.01 & 0.00034* \\ 
        \midrule
        All Scenarios & 26.52 & 64.86 & 5.12 & $<$0.00001* \\
        \bottomrule
    \end{tabular}
    }
\caption{ Fisher's Exact Test performed per task and cumulatively. Asterisks denote statistical significance.}
    \label{tab:accuracy}
\end{table}

\subsection{Satisfaction}

\begin{figure}[t]
\centering
\begin{subfigure}{0.5\linewidth}
        \centering
        \includegraphics[height=1in]{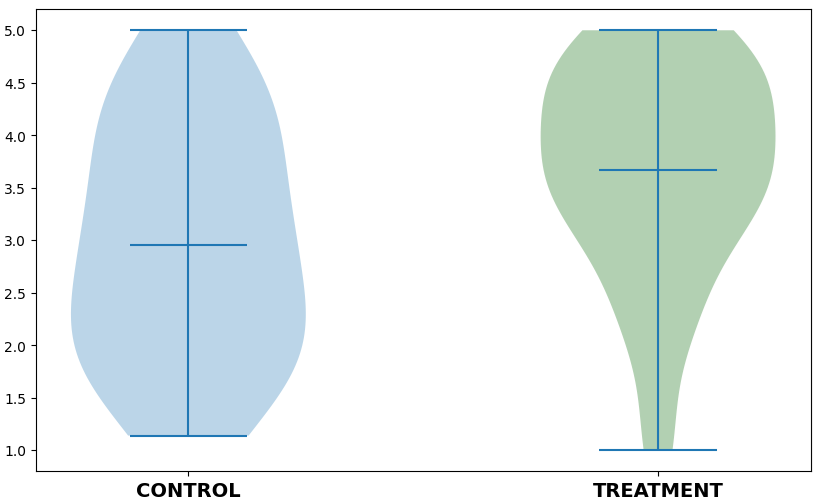}
        \caption{Satisfaction}
        \label{fig:satisfaction}
    \end{subfigure}%
    ~ 
    \begin{subfigure}{0.5\linewidth}
        \centering
        \includegraphics[height=1in]{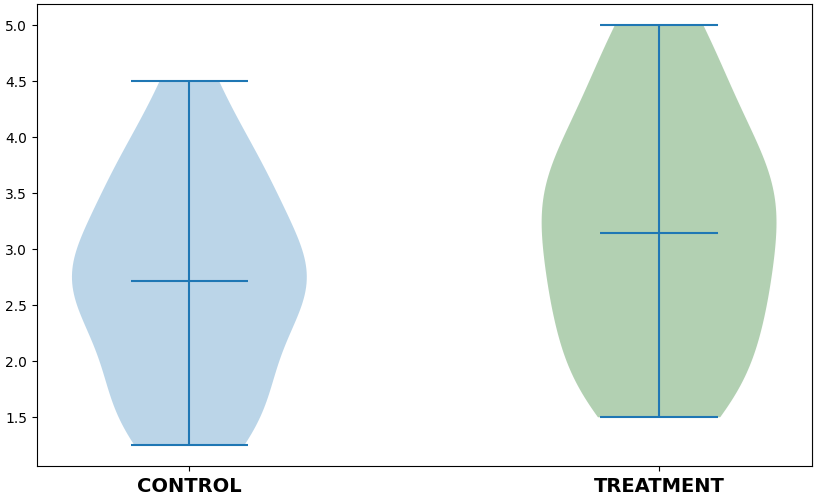}
        \caption{Trust}
        \label{fig:trust}
    \end{subfigure}
    \begin{subfigure}{0.5\linewidth}
        \centering
        \includegraphics[height=1in]{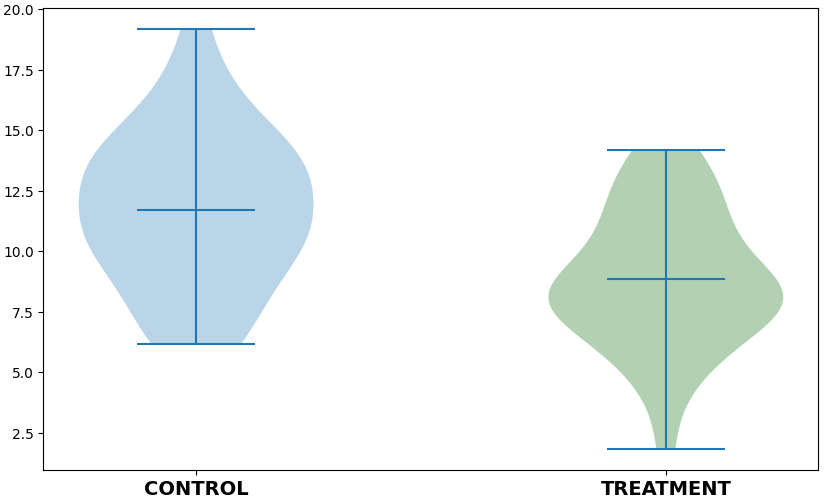}
        \caption{Cognitive Load Index}
        \label{fig:tli}
    \end{subfigure}
\caption{Graphs depicting the distribution of average survey scores assigned by participants for User Satisfaction, Perceived User Trust and Perceived Cognitive Load}
\label{fig:violins}
\end{figure}

The satisfaction survey results are shown in Figure~\ref{fig:satisfaction}, which shows the degree of agreement with questions that ask about satisfaction with explanations. 
We aggregate responses across all questions since they are all directionally the same and look at satisfaction with different aspects of the explanation.
Participants in the treatment group were  significantly ($p \approx 0.0036$) more likely to agree with statements pertaining to the satisfaction of explanations.
This can be seen as a greater mass toward the top of the chart in the treatment group---higher number means more agreement.
{\em Hypothesis H2 is supported.}

\subsection{Trust}

The satisfaction survey results are shown in Figure~\ref{fig:trust}.
As with the satisfaction results, a higher mass indicates more agreement with statements about trust; for the one question framed negatively, we flipped the axis in order to aggregate the responses with the other questions.
Participants in the Treatment group are significantly ($p \approx 0.034$) likely to indicate trust (and less wariness) in the system than those in the Control group.
{\em Hypothesis H3 is supported.}

\subsection{Cognitive Load}

The cognitive load survey results are shown in Figure~\ref{fig:tli}.
We analyzed the results using the Raw TLX technique~\cite{byers1989traditional}, a simple unweighted mean of all answers to summarize the results per participant. 
A lower RTLX score indicates lower workload on the participant, which is preferred. 
Participants in the Treatment group reported significantly ($p \approx 9.76\times 10^{-05}$) lower cognitive loads.
{\em Hypothesis H4 is supported.}

\subsection{Task Completion Time}

We recorded the amount of time that participants took to complete each scenario. 
This encompasses the time they spent reading the status, looking at the video, looking at the images, assembling the response from the pre-populated lists, and writing a short free-text description of how they arrived at the answer. 
Participants in the Treatment group took on average 3.55 (Std.Dev. 2.57) minutes to complete each scenario. Participants in the Control group took on average 3.68 (Std.Dev. 3.09) minutes to complete each scenario. 

A t-test indicates that there is no significant difference in completion times between groups ($p < 0.35$). 
Table~\ref{tab:avg_time_taken} breaks out the time per group and per scenario.
Taken along with other results, the explanations in the Treatment Group neither add cognitive load nor add cognitive processing time.
Three of four scenarios result in less time, though not significantly so. 
One scenario took longer on average; this scenario was also significantly harder (per accuracy results in Table~\ref{tab:accuracy}).

\begin{table}[t]
\ra{1.3}
\resizebox{\columnwidth}{!}{
    \begin{tabular}{lrrrr}
        \toprule
        \multirow{2}{*}{\textbf{Scenario}} & \multicolumn{2}{c}{\textbf{ Mean Time (Std.Dev)}} & \multirow{2}{*}{\textbf{t-statistic}} & \multirow{2}{*}{\textbf{p-value}}\\ 
        \cmidrule{2-3}
        & \textbf{Control} & \textbf{Treatment} & &\\
        \midrule
        Remove Non-Essential Object & 3.80 (2.16) & 4.37 (2.23) & 1.07 & 0.86 \\ 
        Move Essential Object & 3.75 (2.57) & 3.27 (2.93) & -0.72 & 0.24 \\ 
        Obstruct Essential Object & 4.02 (4.78) & 3.65 (2.79) & -0.39 & 0.35 \\ 
        Remove Essential Object & 3.16 (1.88) & 2.91 (1.97) & -0.53 & 0.29 \\ 
        \midrule
        All Scenarios & 3.68 (3.09) & 3.55 (2.57) & -0.38 & 0.35 \\
        \bottomrule
    \end{tabular}
    }
\caption{ Table displaying the average time taken (in minutes) by each group to complete tasks.}
    \label{tab:avg_time_taken}
\end{table}

\section{Discussion}

We overfit our policy model for experimental purposes because we require an agent that will fail in a relatively simple environment. 
It is always the case that an RL agent can fail due to real-world complexity, even when trained with robustness in mind.
Even if the agent does not fail, the user may find the agent to not be aligned to their preferred way of completing a task.
Regardless, the forward and reverse world models must be able to robustly predict state dynamics transitions from parts of the state space that the agent might not commonly visit during training.

The RWM generates predictions of states from within the distribution of what the FWM and agent have trained upon, since FWM and RWM train on the same data. 
The RWM generated suggestions are hence capable of providing insights into the agent's expectations that would be difficult to approximate for an explainer that isn't internal to the agent and wasn't trained alongside the policy. 
This, however, also ties the RWM's capabilities to how well FWM and agent have been trained. 
To create a more robust RWM that can provide good predictions if the counterfactuals are very different from states that the agent would have routinely explored during training, one might need off-policy exploration strategies such as \cite{eysenbach2018diversityneedlearningskills,hafner2019learninglatentdynamicsplanning} .
In general, World Models in RL agents are more likely to catastrophically forget state transition dynamics that are not  directly relevant to the policy construction~\cite{Balloch2024Dissertation}.

\section{Conclusions}

Explainable RL agents for non-AI experts presents a significant challenge because the user cannot change the agent or re-train the policy if it doesn't perform as expected.
To explain why an RL agent chose a particular action over another, counterfactual, action, we generate a  state in which the agent would have chosen the user's desired action.
To do this, we extend the RL agent with a reverse world model that can predict and generate counterfactual states that preceded the current state, instead of the more common prediction and generation of future counterfactual states.

Our human participant study demonstrates that generating the prior counterfactual to the desired action can significantly improve users' abilities to identify the environmental cause of agent failure.
It also improves user satisfaction and trust while reducing cognitive load.

The policy encodes the agent's understanding of how environmental observations map to action that optimizes for future task reward. 
Improving the user's understanding of the agent's policy with respect to the environment helps them update their mental model of the agent.
This in turn helps them identify the cause of failures or the cause of mis-alignment between user and agent policy.
Finally, while not directly studied in the scope of this paper, it also potentially enables the user to correct or align agent behavior through deliberate alterations of the environment.

\begin{figure}[]
    \centering
    \setlength{\fboxsep}{0pt}%
\setlength{\fboxrule}{1pt}%
\fbox{\includegraphics[width=0.60\columnwidth]{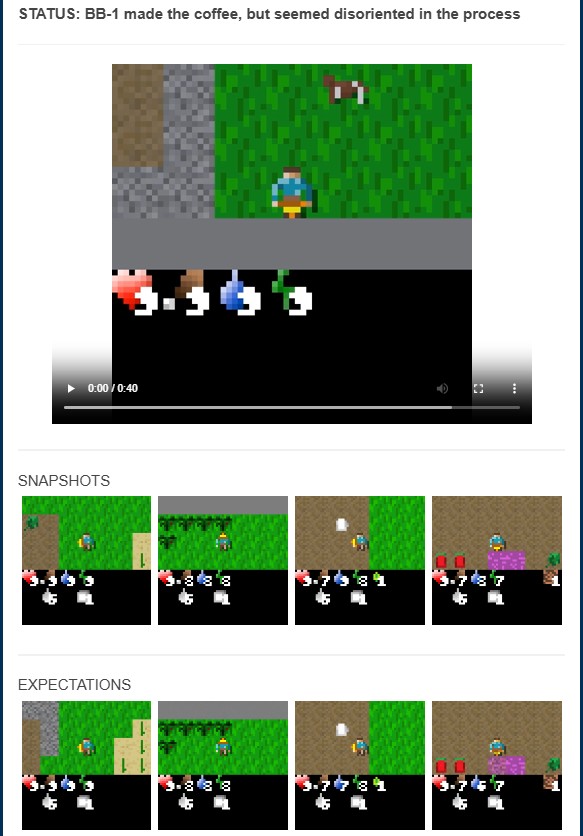}}
    \caption{Screenshot of a scenario as presented to participants consisting of an execution video, a list of images depicting the true snapshots and a list of images depicting the RWM's expectations.}
    \label{fig:qualtrics-question}
\end{figure}

\begin{figure}[h!]
    \centering
        \setlength{\fboxsep}{0pt}%
\setlength{\fboxrule}{1pt}%
\fbox{\includegraphics[width=0.60\columnwidth]{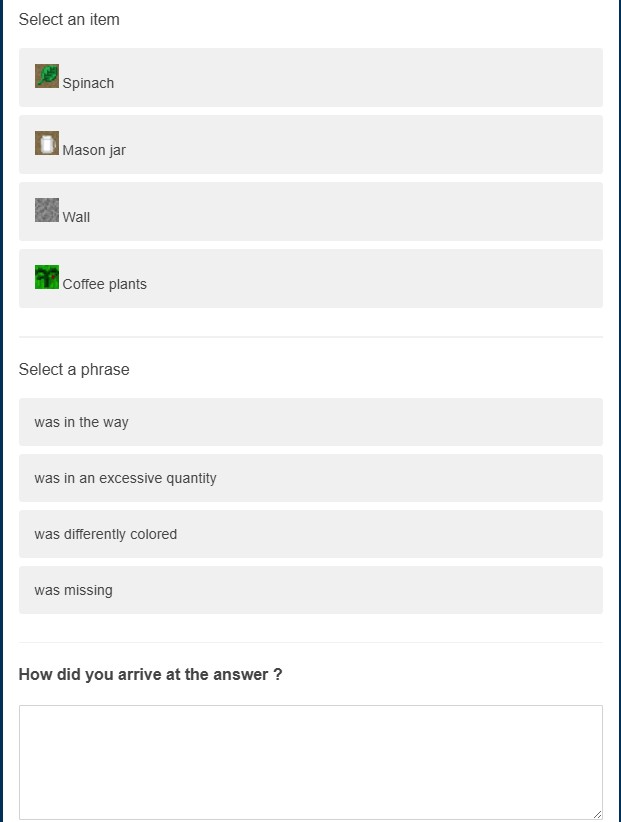}}
    \caption{Screenshot of the interface where participants assemble the cause of the agent's problem from pre-populated lists.}
    \label{fig:qualtrics-answer}
\end{figure}

\section{Appendix}

The core human participant study materials are presented in Figures~\ref{fig:qualtrics-question} and \ref{fig:qualtrics-answer}.

\bibliographystyle{named}
\bibliography{ijcai25}

\end{document}